\documentclass{article}

\usepackage{microtype}
\usepackage{graphicx}
\usepackage{subfigure}
\usepackage{booktabs} 
\usepackage{amsmath}
\usepackage{listings}
\usepackage{xcolor}
\usepackage{subcaption}
\usepackage{amssymb}
\usepackage{hyperref}

\usepackage[accepted]{mlsys2024style/mlsys2024}

\mlsystitlerunning{Flex Attention}

\definecolor{codegreen}{rgb}{0,0.6,0}
\definecolor{codegray}{rgb}{0.5,0.5,0.5}
\definecolor{codepurple}{rgb}{0.58,0,0.82}
\definecolor{backcolour}{rgb}{0.95,0.95,0.92}

\lstdefinestyle{mystyle}{
    commentstyle=\color{codegreen},
    keywordstyle=\color{magenta},
    numberstyle=\tiny\color{codegray},
    stringstyle=\color{codepurple},
    basicstyle=\ttfamily\footnotesize,
    breakatwhitespace=false,         
    breaklines=true,                 
    captionpos=b,                    
    keepspaces=true,                                                 
    showspaces=false,                
    showstringspaces=false,
    showtabs=false,                  
    tabsize=2
}
\begin{document}

\twocolumn[
\mlsystitle{Flex Attention: a Programming Model for Generating \\ Optimized Attention Kernels}



\mlsyssetsymbol{equal}{*}

\begin{mlsysauthorlist}
\mlsysauthor{Juechu Dong}{equal,goo,to}
\mlsysauthor{Boyuan Feng}{equal,goo}
\mlsysauthor{Driss Guessous}{equal,goo}
\mlsysauthor{Yanbo Liang}{equal,goo}
\mlsysauthor{Horace He}{equal,goo}
\end{mlsysauthorlist}

\mlsysaffiliation{goo}{Meta Platforms, Menlo Park, California, USA}
\mlsysaffiliation{to}{Computer Science and Engineering Department, University of Michigan-Ann Arbor, Ann Arbor, MI, USA (work done while at Meta)}

\mlsyscorrespondingauthor{Horace He}{chilli@meta.com}

\mlsyskeywords{Machine Learning, MLSys}

\vskip 0.3in

\begin{abstract}
Over the past 7 years, attention has become one of the most important primitives in deep learning. The primary approach to optimize attention is FlashAttention, which fuses the operation together, drastically improving both the runtime and the memory consumption. However, the importance of FlashAttention combined with its monolithic nature poses a problem for researchers aiming to try new attention variants --- a ``software lottery". 
This problem is exacerbated by the difficulty of writing efficient fused attention kernels, resisting traditional compiler-based approaches. 
We introduce \textit{FlexAttention}, a novel compiler-driven programming model that allows implementing the majority of attention variants in a few lines of idiomatic PyTorch code. We demonstrate that many existing attention variants (e.g. Alibi, Document Masking, PagedAttention, etc.) can be implemented via \textit{FlexAttention}, and that we achieve competitive performance compared to these handwritten kernels. Finally, we demonstrate how \textit{FlexAttention} allows for easy \textit{composition} of attention variants, solving the combinatorial explosion of attention variants.
\end{abstract} 
]



\printAffiliationsAndNotice{\mlsysEqualContribution} 


\section{Introduction}

High-performance self-attention kernels, at the heart of Transformers~\cite{transformer}, have become one of the most important building blocks in deep learning. FlashAttention~\cite{flashattention, flashattention2}, the standard kernel strategy for attention, fuses the tensor operations in self-attention into one kernel, drastically improving the runtime and memory consumption. These optimizations are essential in enabling efficient training and decoding, particularly for long sequences. 
Unfortunately, this performance comes at the cost of flexibility. In exchange for the performance and memory benefits, users are restricted to only the handful of popular attention variants supported. 

However, new attention mechanisms continue to be a focus of significant research. The goal of these changes vary widely, including reductions in computational complexity (Sliding Window Attention ~\cite{slidingwindow}), improvements to training stability (Softcapping ~\cite{gemma2}), ability to work with sequences of differing lengths (Document Masking), better length extrapolation (ALiBI~\cite{alibi}), applying attention to other domains such as images (Neighborhood Attention \cite{NAdeliated}), modifications for better inference throughput (PagedAttention~\cite{PagedAttn}), etc. Furthermore, users often want combinations of these (such as Sliding Window Attention combined with ALiBI), leading to a combinatorial explosion of possible attention kernels. These modifications are not merely speculative either -- many of the most popular Large Language Models (LLMs) released use these variants, such as Sliding Window Attention in Mistral-7B, Softcapping in Gemma-2, or ALiBI in MPT-7B.

The importance of FlashAttention combined with its limited ability to support all attention variants poses a problem for researchers aiming to try new attention variants — a "software lottery" of sorts. If a particular attention variant is not supported by the existing kernels, further exploration can be hindered by slow runtime and memory limitations.

This problem is exacerbated by the difficulty of automatically generating fused attention kernels, resisting traditional compiler-based approaches. In addition to all the standard difficulties with generating efficient matrix multiplication-based primitives on accelerators, there also exist several algebraic rewrites that are difficult for compilers to perform. For example, although Mirage~\cite{mirage} is able to generate attention forward from primitive operations, it misses key components for practical usage of attention such as safe softmax or the backwards pass. Furthermore, many variants of attention require block sparsity, a further difficulty for traditional compiler-based approaches.

\subsection{Our Approach}
We present \textit{FlexAttention}, a novel compiler-driven programming model that allows implementing the majority of attention variants in a few lines of idiomatic PyTorch code. 

\textbf{Flexible Programming Model. } Instead of a constrained interface designed with specific attention variants in mind, \textit{FlexAttention} is a general programming model for attention that can be used to define many different attention flavors.
We observe that many attention variants can be defined as a score modification applied on the intermediate score matrix before conducting softmax (Eq. \ref{eq:attn}). Additionally, a good portion of these modifications take the form of masks, which set part of the score matrix to {\tt -inf}. 

{\small
\begin{align}
    \text{Attention}(\mathbf{Q}, \mathbf{K}, \mathbf{V}) &= \text{softmax}(\frac{\mathbf{Q}\mathbf{K}^T}{\sqrt{d_k}})\mathbf{V} \nonumber \\ 
    \text{\textcolor{blue}{Flex}Attention}(\mathbf{Q}, \mathbf{K}, \mathbf{V}) &= \text{softmax}(\text{\textcolor{blue}{mod}}(\frac{\mathbf{Q}\mathbf{K}^T}{\sqrt{d_k}}))\mathbf{V} 
\label{eq:attn}
\end{align}
}%
With these insights, we build \textit{FlexAttention}, which takes a score modification callable ({\tt score\_mod}) and an attention mask callable ({\tt mask\_mod}) in addition to tensor inputs. It enables users to implement a new attention variant by defining how to update scores or calculate boolean masks based on positional information. We demonstrate that many existing attention variants (e.g. Alibi, Document Masking, PagedAttention, etc.) can be implemented via \textit{FlexAttention}. In addition, our programming model allows for easy composition of attention variants via nested {\tt score\_mod}s and boolean operations between {\tt mask\_mod}s, solving the combinatorial explosion 
of attention variants. 

\textbf{Handwritten Template-Based Generation}
During PyTorch compilation, {\tt score\_mod} and {\tt mask\_mod} are lowered and codegened into the main loop of a handwritten attention kernel. This abstraction allows us to leverage the performance of hand-written kernels behind the scenes, while still providing significant flexibility in the frontend.
The {\tt torch.compile} lowering framework is natively compatible with PyTorch and offers strong flexibility to handle diverse {\tt score\_mod} and {\tt mask\_mod} defined by the users. Additionally it also supports automatic backward pass generation via {\tt torch.autograd}. 
We build forward and backward graphs for {\tt score\_mod} and {\tt mask\_mod}, perform operator fusion and produce kernel code blocks to be integrated with attention kernel templates. Compute buffers are pre-allocated for inputs, outputs and saved intermediate values.

\textbf{Block Sparsity Optimization.} In addition to its semantic changes, masking also introduces sparsity.
To take advantage of this, \textit{FlexAttention} leverages block sparsity and employs a pre-computed {\tt BlockMask}.
{\tt BlockMask} is a small matrix that tracks block-level sparsity on wether a tiled score matrix block is fully masked out.
With the help of {\tt torch.vmap}, our {\tt create\_block\_mask} utility automatically generates \texttt{BlockMask} from user defined {\tt mask\_mod}.
The {\tt BlockMask} unlocks the opportunity to save the work for fully-masked score matrix blocks without loading a large elementwise attention mask. 
For partially masked score matrix blocks, we can still exploit \texttt{mask\_mod} to elementwisely mask out score scalars, thus maintaining the semantics.
Additionally, {\tt BlockMask} is implemented as a index vector which can also serve as the mapping used in PagedAttention\cite{PagedAttn} (\autoref{sec:paged_attn}).

We evaluate \textit{FlexAttention} on 7 popular attention variants and compare its performance against today's most commonly used infrastructures: PyTorch SPDA, and handwritten kernels from FlashAttention (FAv2 \cite{flashattention2}, FAv3 \cite{FA3} and FAKV\cite{flashdecoding}). We show that \textit{FlexAttention} delivers 0.68x-1.43x the performance of FAv2 and 0.93x-1.45x of FAKV for decoding on conventional attention variants supported by FlashAttention. \textit{FlexAttention} works well with existing ML infrastructure and improves end-to-end performance for inference by 2.04x in gpt-fast for 16k context length and training by 2.4x in torchtune. Additionally, we show that \textit{FlexAttention} supports paged attention at negligible overhead. 
\section{Background}
\subsection{Attention Variants}
The attention mechanism plays a critical role at the core of Transformers.
Each attention layer takes three input tensors: a query $\mathbf{Q} \in \mathbb{R}^{B \times H \times Q\_LEN \times D}$, and a key and a value $\mathbf{K},\mathbf{V} \in \mathbb{R}^{B \times H \times KV\_LEN \times D}$, where $B$ denotes the batch size, $H$ is the number of attention heads, $D$ is the feature dimension, and $Q\_LEN$ and $KV\_LEN$ represents the query and key-value lengths, respectively. The attention mechanism first computes a score matrix $\mathbf{S} \in \mathbb{R}^{B \times H \times Q\_LEN \times KV\_LEN}$, which encodes context information by attending each query token to every key token.

\begin{equation}
    \mathbf{S} = \text{softmax}(\frac{\mathbf{Q}\mathbf{K}^T}{\sqrt{d_k}})
\end{equation}

Next, attention computes the output as $\mathbf{S}\mathbf{V} \in \mathbb{R}^{B\times H \times Q\_LEN \times D}$, weighting the value features by the score matrix. 

Building on this foundational algorithm, ML researchers are exploring methods to enhance it by modifying the score matrix $\mathbf{S}$ for more effective and efficient context extraction.  For example, neighborhood \cite{NAdeliated} and sliding window~\cite{slidingwindow} attention attends each query token only to its neighboring key tokens to reduce computational and memory complexity for large images and long sequences. Softcapping~\cite{gemma2} adds a tanh layer to prevent excessive growth of logits. Alibi~\cite{alibi} embeds an elementwise bias in the score matrix that penalizes distant tokens to allow models trained on short inputs to perform on long prompts. This large design space motivates efficient compiler designs for facilitating the training and inference of attention variants.

\subsection{State-of-the-art Attention Implementations}
\label{sec:sota_attn}

Many kernels have been manually implemented to efficiently support attention. Among these, the IO-aware FlashAttention \cite{flashattention2} has become one of the most widely adopted solutions. FlashAttention significantly reduces memory access and achieves substantial speedups. 
Specifically, it avoids materializing the large score matrix $\mathbf{S}$ and computes it on the fly.
Recently, Flash Attention v3 has been released to further accelerate attention by leveraging advanced hardware features and manually tuning the performance.
While flash attention delivers state-of-the-art performance, it is specifically tailored towards a limited selection of attention variants (\autoref{tab:evaluation:mods}).
One recent work FlashMask \cite{wang2024flashmask} extends Flash Attention with a column-wise sparse representation to support more mask designs.
However, it still lacks of flexibility in terms of score modifications and adds large overhead for complex masks.
Given that attention variants exhibit diverse computational characteristics such as sparsity and locality, significant manual effort is required to adapt existing attention implementations to support numerous attention variants.
This lack of flexibility and efficient kernels has become a barrier for machine learning researchers seeking to explore novel attention variants.

\subsection{Machine Learning Compilers}
Many compilers have been built to accelerate machine learning workloads.
torch.compile \cite{pytorch2} uses TorchDynamo to capture computation graph from an arbitrary code and TorchInductor to optimize the graph.
TVM \cite{TVM} allows users to describe the computation of a kernel and generate optimized code.
Mirage \cite{mirage} utilizes $\mu$graph to explore operator-level optimization opportunities. 
However, these machine learning compilers fail to deliver a satisfactory performance on attention variants due to the special computing patterns in attention.
First, attention requires fusing two matrix-matrix multiplications (i.e., $QK^T$ and $SV$) while existing machine learning compilers usually focuses on optimizing one matrix-matrix multiplications.
Second, as shown in Flash Attention \cite{flashattention2}, online softmax significantly reduces memory access and improve performance. Since online softmax is tailored towards attention designs, it is not well supported by existing general machine learning compilers.
To the best of our knowledge, we are the first to compile arbitrary attention variants while delivering state-of-the-art performance.


\begin{figure*}
    \centering
    \includegraphics[width=\linewidth]{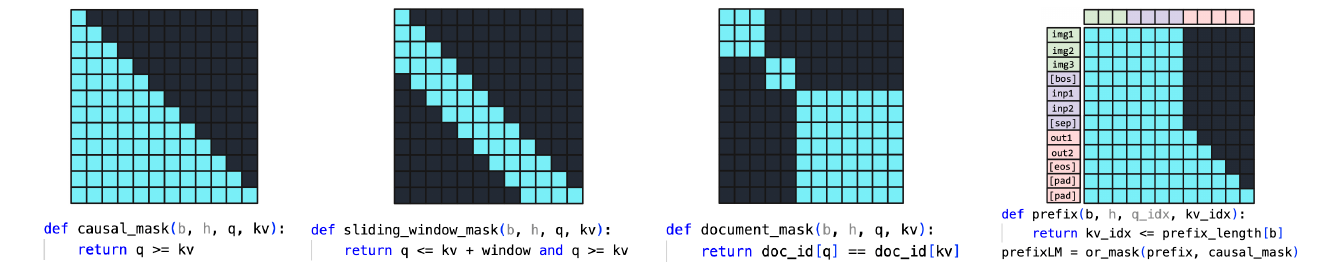}
    \caption{Attention variants and \texttt{mask\_mod} examples, including causal mask, sliding window mask, document mask, and prefixLM mask.}
    \label{fig:mask_mod}
\end{figure*}

\section{Front-end Design and Implementation}
\label{sec:API}

In this section, we propose a unified abstraction of diverse attention variants.
This abstraction enables programmers to easily express attention semantics without worrying about implementation details and kernel performance.

\subsection{Unified Abstraction}
\label{sec:score_mod}
Despite the numerous attention variants designed by machine learning researchers, we unify these variants into two commonly shared patterns. The first pattern masks out features from certain tokens to manage contextual relationships. For example, a causal mask forces a model to predict based on previous tokens and ignore future tokens. A sliding window mask~\cite{slidingwindow} forces a model to concentrate on nearby tokens for the local context. 

The second pattern further adjusts the attention score among tokens in a fine-grained style. For example, \textit{Alibi}~\cite{alibi} adjusts the attention score based on the relative position of two tokens, which helps the model to better understand the context and handle long sequences. It also adds bias to attention scores based on head dimension, which specializes heads on different types of information.

With these insights, we design a unified abstraction including a score modification callable ({\tt score\_mod}) and an attention mask callable ({\tt mask\_mod}).
It enables users to build a new attention variant by defining how to update scores or calculate boolean masks based on positional information in idiomatic PyTorch code.
{\small
\begin{lstlisting}[language=Python]
def mask_mod(batch_idx: int, head_idx: int, q_idx: int, kv_idx: int) -> bool

def score_mod(score: T, batch_idx: int, head_idx: int, q_idx: int, kv_idx: int) -> T
\end{lstlisting}
}

Specifically, consider a score matrix $\mathbf{S} \in \mathbb{R}^{B\times H\times M \times N}$. 
\texttt{mask\_mod} takes the input position and specifies whether the corresponding score scalar is set to {\tt -inf}.
\texttt{score\_mod} additionally takes the score scalar of arbitrary type T (e.g., \texttt{bfloat16} or \texttt{float32}) and applies updated to the score based on its position.

\paragraph{Mapping to concrete examples}
Our unified abstraction captures the two important patterns in attention variants and enables automated compiler optimizations.
\autoref{fig:mask_mod} demonstrates 4 example attention variants, 
each of which has been widely studied and optimized. However, there is a lack of a unified optimization approach that applies to all of them. 
Instead, our abstraction provides full flexibility in specifying masking and modification to the score matrix, enabling machine learning researchers to easily explore novel attention variants. 

As we discussed earlier, a \texttt{causal mask} forces a token's query feature to only attend to key and value features from previous tokens. This leads to a simple expression of $q\_idx \ge kv\_idx$. Since the causal mask is not related to the batch and head dimension, the \texttt{mask\_mod} does not involve $b$ and $h$ inputs.

A \texttt{sliding window mask} requires a token to only attend to nearby tokens within a certain sliding window size. We can easily describe such requirement as $q\_idx - kv\_idx \le window$.

A \texttt{document mask} trains multiple documents simultaneously and requires a token to only attend to other tokens in the same document. We can record `document\_id` as a mapping from the token index to the document index, and keep the score only if $document\_id[q\_idx] == document\_id[kv\_idx]$.

A \texttt{alibi bias} modifies each score scalar by adding a bias scalar, which scales with the distance between the query index and the key value index. We can easily describe it as $score + alibi\_bias[h] * (q\_idx - kv\_idx)$. Note that we can easily support other score modifications such as scaling the score scalar or even non-linear transformations.

\paragraph{Why \texttt{mask\_mod}?}
\texttt{mask\_mod} is a special case of \texttt{score\_mod} and any \texttt{mask\_mod} can be easily converted to a \texttt{score\_mod} semantically.
However, we still observe two fundamental reasons for distinguishing these two APIs.
First, \texttt{score\_mod} applies expensive modification to every score scalar. When converting \texttt{mask\_mod} to \texttt{score\_mod}, we multiply each score with $1$ or $0$, leading to extra overhead.
Second, comparing with \texttt{score\_mod}, \texttt{mask\_mod} provides extra semantic information that certain score computations can be skipped. 
We exploit this information to further accelerate attention computation, as detailed in \autoref{sec:block_sparsity}.

\subsection{Logical Fusion for Composability}
\label{sec:logical_fusion}
As many attention variants have been designed, one recent trend is to compose existing attention variants to further extract semantic information.
We support the logical fusion to enable the composability of mask designs via \texttt{and\_mask} and \texttt{or\_mask}.
By taking two \texttt{mask\_mod} functions, we can automatically compose them by applying elementwise logical operations.
The generated \texttt{mask\_mod} can be further combined with other \texttt{mask\_mod} which significantly mitigates the programmability burden when designing attention variants.

\autoref{fig:mask_mod} shows an example of logical fusion. \texttt{PrefixLM} \cite{T5} performs full bidirectional attention on prefix inputs and causal attention on the rest.
Instead of complex conditional branches, we can build a simple \texttt{prefix mask} and compose it with a causal mask via \texttt{or\_mask}.

\section{Backend Design and Implementation}
In this section, we propose a backend design that compiles attention variants to efficient kernels.

\subsection{Template-based Lowering}
We build a template-based lowering system that generates optimized GPU kernels, combining the flexibility with state-of-the-art attention optimizations.
The key innovation of our approach lies in recognizing that attention variants primarily differ in their pointwise score modifications, allowing us to templatize the common computational patterns.
To this end, we build template-based lowering to capture the common patterns and apply \texttt{score\_mod} elementwisely on score matrix.
We will present BlockSparsity design on exploiting sparsity from \texttt{mask\_mod} in \autoref{sec:block_sparsity}.

Template-based lowering first exploits TorchDynamo \cite{pytorch2} to capture the computation graph of \texttt{score\_mod} and \texttt{mask\_mod}.
As shown in \autoref{fig:mask_mod}, these two functions are usually lightweight in terms of computation and memory access, and could be fused with other computation in attention.
Then, we build a highly optimized triton template for high performance.
The Triton kernel design adopts established optimization techniques for fused attention kernels, including online softmax, careful GPU occupancy management, efficient memory handling via partitioning and broadcasting, and specialized support for grouped query attention (GQA) \cite{GQA}.

\begin{figure}[h]
    \centering
    \includegraphics[width=\linewidth]{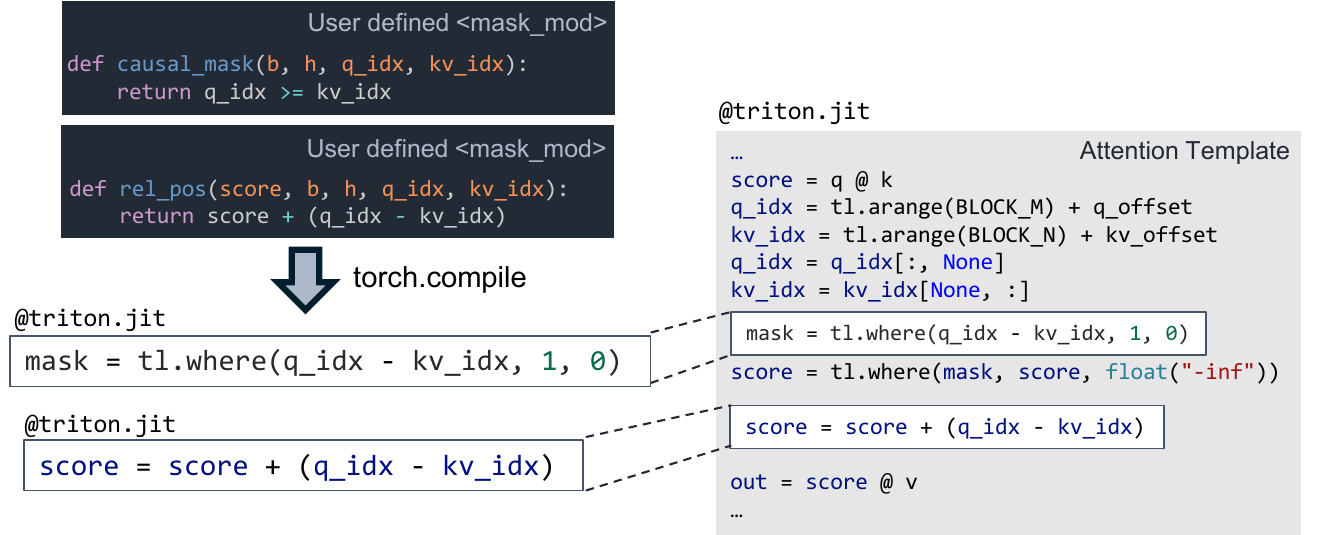}
    \caption{Compilation from user-defined attention patterns to optimized Triton kernels. \textbf{Top:} User-defined PyTorch functions specify custom masking (\texttt{causal\_mask}) and score modification (\texttt{rel\_pos}) operations. \textbf{Bottom:} \texttt{torch.compile} captures and lowers these operations into Triton primitives. \textbf{Right:} The resulting Triton code blocks are integrated into our attention template, which handles core attention computations while preserving the user-defined modifications.}
    \label{fig:lowering}
\end{figure}

We capture these advanced techniques in three handwritten attention kernel templates (forward, backward, and decoding). The templates are designed to accept custom score modification code blocks, which are generated from captured score\_mod and mask\_mod operations using torch.compile. TorchInductor translates these subgraphs into Triton code, dynamically injecting both forward and backward score modification operations into the predefined templates at runtime. This approach results in attention kernels that closely match the performance of manually optimized, variant-specific implementations. Figure~\ref{fig:lowering} illustrates our lowering pipeline, showing how user-defined PyTorch operations are translated into optimized Triton code and integrated into our attention templates.

\subsection{Block Sparsity}
\label{sec:block_sparsity}

Attention variants usually show high sparsity such as 50\% sparsity from causal mask and significantly higher sparsity from sliding window mask.
One natural question is, 

\textit{could our compiler exploit such sparsity for acceleration?}

One naive approach is to check during runtime whether a position is masked out and skip the computation.
However, this adds a large runtime overhead by iterating through all scalars even if it is masked out.
Another approach is to pre-compute a mask tensor of shape $B\times H\times Q\_LEN \times KV\_LEN$ and compute a score scalar according to the mask tensor.
However, this mask tensor adds significant memory overhead, which contradicts the principle of flash attention that avoids realizing the score matrix.

\paragraph{BlockMask}
FlexAttention implements a BlockMask data structure to exploit the sparsity while adding only negligible memory overhead.
We pre-compute a block mask during compilation time to remove the runtime overhead.
BlockMask first splits the score matrix into blocks along the Q\_LEN and KV\_LEN dimensions.
Then, BlockMask specifies a block as non-computed if all score scalars in it are masked as {\tt -inf}.
By recording sparsity at the block level, we do not realize a large sparsity matrix, significantly reducing memory overhead.

Concretely, BlockMask contains two tensors, a \texttt{kv\_num\_block} of shape $B\times H \times Num\_Row$ and a \texttt{kv\_indices} of shape $B\times H \times Num\_Row \times Num\_Col$.
Here, $Num\_Row$ and $Num\_Col$ are the number of blocks in each row and column along the query and key value dimension, respectively.
\texttt{kv\_num\_block} stores the number of non-zero blocks for each row and \texttt{kv\_indices} stores the indices of these non-zero blocks.
By accessing \texttt{kv\_num\_block}, \textit{FlexAttention} skips masked blocks and achieves proportional performance speedup.

\begin{figure}[t]
    \centering
    \includegraphics[width=\linewidth]{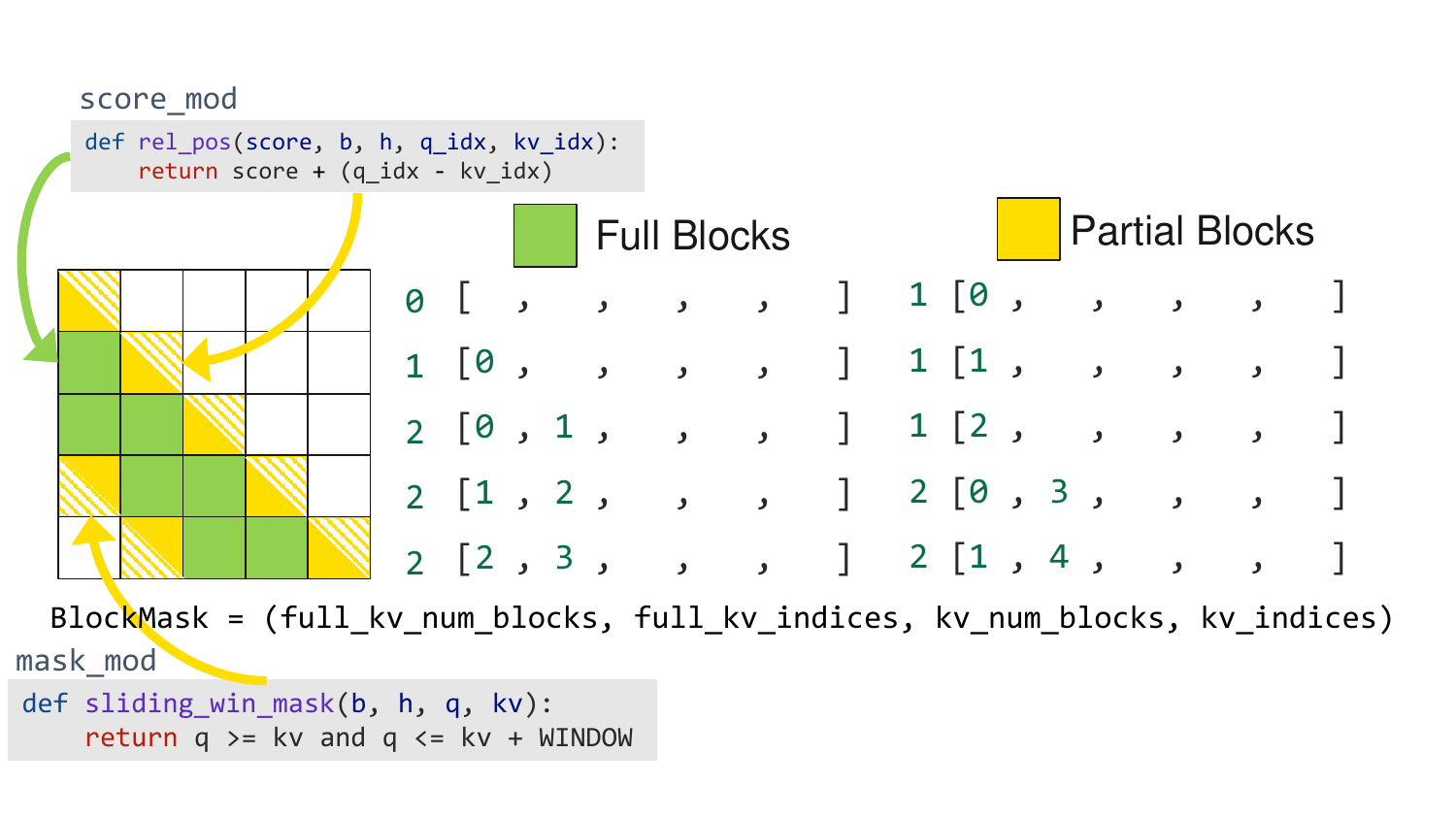}
    \caption{BlockMask for Sliding Window Attention. \textbf{Left:} The score matrix is divided into fully visible blocks (green), partially masked blocks (yellow), and oblivious blocks (white). \textbf{Right:} {\tt BlockMask} encoded as an index matrix plus block count per row. \textbf{Top: }{\tt score\_mod} is applied to full \& partial blocks. \textbf{Bottom: }{\tt mask\_mod} is applied to partial blocks. }
    \label{fig:BLockMask}
\end{figure}

\paragraph{Full Block Optimization}
We split BlockMask into full blocks and partial blocks to further improve the performance.
Our key idea is to minimize runtime overhead caused by applying \texttt{mask\_mod}, skipping it whenever possible.
We identify two types of blocks in terms of sparsity.
\begin{enumerate}
    \item \underline{Partial blocks} where some score scalars are masked as {\tt -inf}, necessitating the application of \texttt{mask\_mod} elementwise at runtime.
    \item \underline{Full blocks} where no score scalar is masked, allowing us to skip {\tt mask\_mod} and apply only {\tt score\_mod}. 
\end{enumerate}
This optimization yields approximately a 15\% performance improvement for common patterns such as causal masks.
\autoref{fig:BLockMask} illustrates the BlockMask for sliding window attention. 
The \texttt{sliding\_win\_mask} is applied only to partial blocks while the \texttt{rel\_pos} score modification is applied elementwise on both full and partial blocks.

\paragraph{BlockMask Guided Indirect Memory Access}

Exploiting sparsity in attention shows significant performance speedup.
Existing solutions such as FlashAttention iterate through the $KV\_LEN$ dimension and manually specifies the start and end index of iteration based on attention variants. This leads to a significant programmability burden.
Moreover, it requires extra efforts to ensure that attention scores in the final block are properly masked if the query token index is less than the key token index.

FlexAttention utilize the BlockMask information to automatically exploit the sparsity in attention variants.
It first adjusts the workload in each GPU block according to the \texttt{kv\_num\_block}, which specifies the number of score matrix blocks that are not masked out.
Then, it uses the KV Indices to map to the next block to be processed.
Here, we utilizes an indirect memory access strategy since the indices need not point to contiguous tokens in the sequence.
This provides flexibility in compiling various attention patterns, such as sliding window attention, local-global attention, or custom sparse patterns, without modifying kernel.

\begin{figure}[!h]
    \centering
    \includegraphics[width=0.7\linewidth]{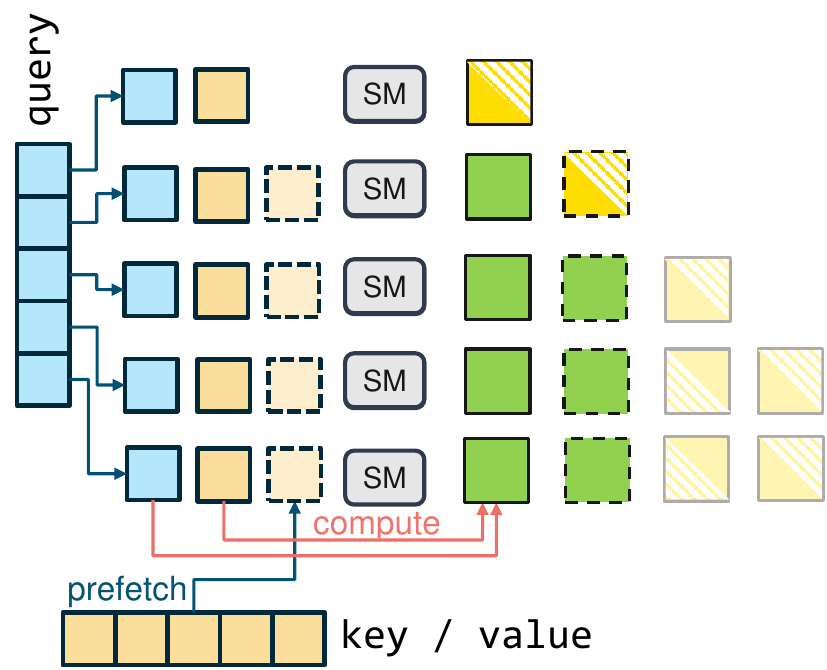}
    \caption{Scheduling full and partial blocks to SM. }
    \label{fig:blockmaskexc}
\end{figure}

\paragraph{Data Prefetching Pipeline}
\textit{FlexAttention} divides the score matrix into tiles along the Q\_LEN dimension to parallelize the computation on multiple Stream Processors(SM). Each SM iterates along the KV\_LEN dimension and processes a row of blocks (Figure \ref{fig:blockmaskexc}).
Special care needs to be taken in pipelining the data fetch and compute for these blocks.
When the first score block is being computed from the first kv tile, the second kv tile (shown in dashed borders) is being pre-fetched from HBM to SRAM.
Our BlockMask access strategy removes conditional branch on checking whether a score scalar is masked out, thus allowing efficient pipelining and hiding data access latency between iterations.

\paragraph{Overhead Analysis}
BlockMask representation achieves memory efficiency through a careful balance of granularity and overhead. Although we store auxiliary tensors to encode the sparsity pattern, their memory footprint scales with $O(\lceil Q\_LEN/BS \rceil \times \lceil KV\_LEN/BS \rceil)$, where $BS$ is the block size (=128 by default).
This is significantly less than the $O(M \times N)$ required for the full score matrix.

\section{Case Study}

\begin{figure*}[ht]
\centering
     \includegraphics[width=\linewidth]{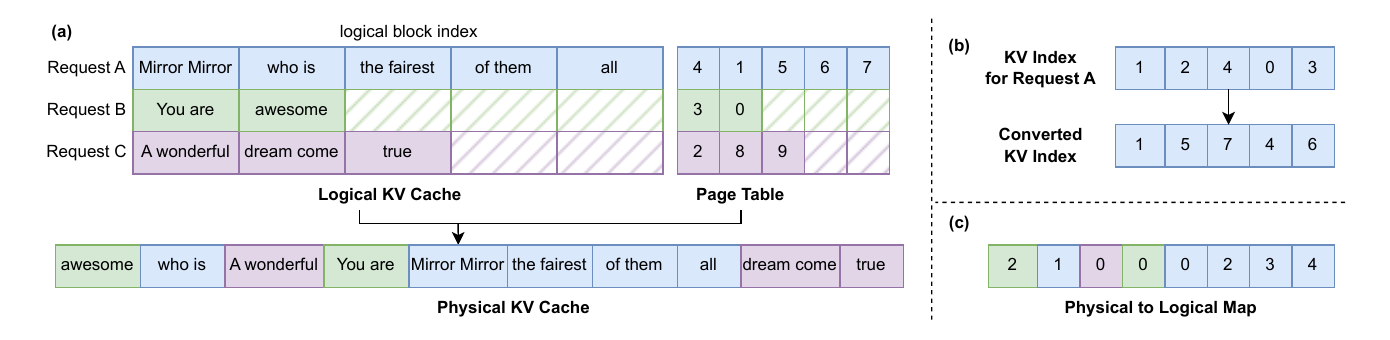}
  \caption{Illustration of \textit{FlexAttention} with paged attention. a) Update physical KV cache with a page table; b) Convert a block mask especially KV Index; c) Physical to logical map for converting \texttt{mask\_mod} and \texttt{score\_mod}.}
  \label{fig:paged_attention}
\end{figure*}

\subsection{Paged Attention Variants}
\label{sec:paged_attn}

PagedAttention \cite{PagedAttn} has been widely deployed for inference on a batch of sentence requests to reduce GPU memory consumption.
However, the current PagedAttention design requires manually rewriting the attention kernel to support page table lookup for its irregular memory accesses.
Additionally, it is tightly coupled with specific attention mask designs, limiting composability with other attention variants. In this subsection, we explain how \textit{FlexAttention} enables paging support for arbitrary attention variants without manually rewriting the kernels, while still delivering high performance and low memory usage.
\footnote{Due to the scope of this paper, we focus on supporting KV cache in GPU memory and leave the memory swapping to host disk as future work.}

\paragraph{Page Table.}
PagedAttention introduces a page table to efficiently manage the KV cache and reduce its memory usage.
As shown in \autoref{fig:paged_attention}(a), a logical KV cache stores the key and value features of each sequence, typically in shape $B\times Max\_len \times D$ where $B$ is the batch size, $Max\_len$ is the maximum context length, and $D$ is the feature dimension.
Since sequence lengths vary significantly between sentences and could be much shorter than $Max\_len$, a substantial portion of the logical KV cache remains unused.
To address this fragmentation issue, PagedAttention instead allocates a physical KV cache of shape $1 \times Max\_token \times D$, where $Max\_token$ is the maximum number of tokens across all sentences combined.
Storing sentences in the physical KV cache reduces fragmentation and enables memory optimizations such as KV block sharing between sentences.

PagedAttention maintains a page table as 2D matrix that maps the batch index and logical KV index to the corresponding physical KV index. Within the attention kernel, one  \texttt{scatter} operation is used efficiently translate indexes using this matrix, while minimizing memory access overhead. During runtime, once a sentence reaches its end token, we clear the corresponding row in the page table and update it with a new request.

\paragraph{Fused indirect memory access.}
Having attention kernel compute on the physical KV cache instead of the logical KV cache, presents several challenges.
First, the page table adds an extra layer of indirection in memory accesses, which existing designs typically handle with manually rewritten CUDA kernels.
Second, the page table imposes a significant programming burden on already complex implementations for supporting diverse attention variants.

\textit{FlexAttention} tackles this problem through the BlockMask conversion, eliminating the need for kernel rewrites.
Our key idea is that since BlockMask already incorporates one layer of indirect memory access to avoid unnecessary computations (\autoref{sec:block_sparsity}), we can merge it with the indirect memory access from the page table to further skip unnecessary memory accesses.
Specifically, as shown in \autoref{fig:paged_attention}(b), given a BlockMask of an attention variant, we take the kv\_index and map the logical block index to the corresponding physical block index according to the page table.
During runtime, \textit{FlexAttention} relies on the converted kv\_index to access tokens for a specific sentence from the physical KV cache.
Note that we keep the kv\_num\_blocks unchanged since paged attention does not change the number of unmasked blocks.

\paragraph{\texttt{mask\_mod} and \texttt{score\_mod} Conversion}
\textit{FlexAttention} supports attention variants with \texttt{mask\_mod} and \texttt{score\_mod}, which utilizes position information (e.g., logical KV index) to specify mask and score modifications.
This position information is changed when using paged attention and operating on the physical KV cache.
One naive approach is to manually rewrite the \texttt{mask\_mod} and \texttt{score\_mod} to cater to such changes on position information.
However, this adds an large programmability burden, considering numerous attention variants and even logical fusion of these variants (\autoref{sec:logical_fusion}).

We automatically compile the \texttt{mask\_mod} and \texttt{score\_mod} to support paged attention.
{\small
\begin{lstlisting}[language=Python]
def converted_mask_mod(batch_idx: int, head_idx: int, q_idx: int, physical_kv_idx: int) -> bool

def converted_score_mod(score: T, batch_idx: int, head_idx: int, q_idx: int, physical_kv_idx: int) -> T
\end{lstlisting}
}
Specifically, we maintain a vector mapping physical block indices to logical block indices using $O(1)$ overhead when updating the page table.
Given the physical KV token index, we can first compute the physical KV block index and offset due to the fixed block size.
Then, we can lookup the corresponding logical KV block index and regenerate the logical KV token index.
Finally, we call the user-provided \texttt{mask\_mod} and \texttt{score\_mod} with the generated logical KV token index.

\begin{figure}
    \centering
    \includegraphics[width=\linewidth]{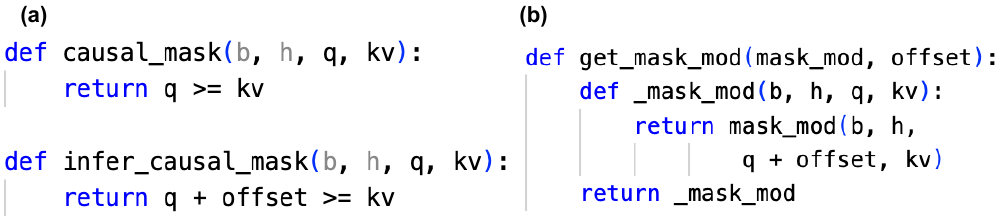}
    \caption{\texttt{mask\_mod} conversion}
    \label{fig:offset}
\end{figure}

\subsection{Modification Conversion for Inference}
In attention variants, the mask and score modifications may change with query indices, which is supported by taking $q\_idx$ in \texttt{mask\_mod} and \texttt{score\_mod}.
However, LLM inference usually iteratively takes one query token at a time and requires additional context information in terms of the number of query tokens has been processed, namely offset.
\autoref{fig:offset}(a) shows the difference between causal mask during training and inference.
To tackle this problem, we automatically convert \texttt{mask\_mod} and \texttt{score\_mod} to their inference counterpart, as shown in \autoref{fig:offset}(b).
Specifically, we provide a decorator that takes the user-defined \texttt{mask\_mod} and an offset, and generate a converted \texttt{mask\_mod} that consumes the offset information.


\section{Evaluation}

\subsection{Experiment Setup}\begin{table*}[!ht]
 \small
    \centering
    \begin{tabular}{|c||c|c|c|c|c|} \hline 
          &  FAv2 & FAv3 ({\tt c1d146c}) & SDPA - cuDNN (v9.1.1)  &  SDPA - mem\_efficient & \textit{FlexAttention} \\ \hline \hline
         noop & native & native & native & native & native\\ \hline 
         causal &  native& native &  native& native & native\\ \hline 
         alibi\_bias&  native& x &  itemized mask& itemized mask & native\\ \hline 
         sliding\_win & native &  native &itemized mask & itemized mask & native \\ \hline
         prefix\_lm&  x& x &  itemized mask& itemized mask & native\\ \hline 
         soft\_cap&  native& x &  x& x & native\\ \hline 
         various lengths &  native& native &  jagged tensors& jagged tensors & native via document\_mask\\ \hline
         neighbor attention & x & x & x & x & native \\ \hline
    \end{tabular}
    \caption{Tested Attention Variants and their Level of Support in FlashAttention, SDPA \& \textit{FlexAttention}}
    \label{tab:evaluation:mods}
\end{table*} 

We evaluate the performance of \textit{FlexAttention} on 7 popular attention modifications (Table. \ref{tab:evaluation:mods}), including the classic {\tt noop} and {\tt causal}, popular positional embedding {\tt alibi}~\cite{alibi}, local attention {\tt sliding\_window} \cite{longformer}, fully visible prefix {\tt prefixLM}, additional tanh layer {\tt soft\_cap} and {\tt document\_mask} for batching input of various lengths. These attention variants are evaluated as Multi-head Attention (MHA) and Grouped Query Attention (GQA).

These attention variants have varying levels of support across our five baselines. \textbf{FlashAttention-v2} ({\tt FAv2})~\cite{flashattention2} is a state-of-the-art, high-performance attention kernel optimized to reduce HBM read/write through kernel fusion, tiling and recomputation. \textbf{FlashAttention-v3}~\cite{FA3} ({\tt FAv3}) is an experimental kernel that leverages new hardware features in Hopper GPUs to further enhance performance. FlashAttention also provides a \textbf{FlashDecoding} ({\tt FAKV}) kernel optimized for inference with kvcache support. Scale dot-product attention (SDPA) is the native PyTorch functional API for attention, supporting\textbf{ math}, \textbf{memory-efficient attention} ({\tt mem\_efficient})~\cite{mem_efficient}, FAv2, and \textbf{cuDNN} backends. SDPA integrates smoothly into the PyTorch framework, enabling optimizations and features such as {\tt NestedTensor}, {\tt torch.compile}, and CUDA graph support. \par

We evaluate end-to-end training and inference performance on LLaMa3 and LLaMa3.1 models~\cite{llama3} using {\tt gpt-fast}~\cite{gpt-fast} and {\tt torchtune}~\cite{torchtune}. {\tt gpt-fast} and {\tt torchtune} are native PyTorch libraries that provide accessible fine-tuning and inference recipes for popular models They rely on {\tt torch.compile} to enable optimizations, including CUDA graphs, kernel fusion and inlining, matmal templates, parameter freezing etc., and use SDPA for attention by default. \par
Our experiments are conducted on Nvidia H100 GPUs with power capped at 650W and memory bandwidth limited to 2.4TB/s, Nvidia A100 GPUs with power capped at 330W, and Nvidia A6000 GPUs.

\subsection{Attention Kernel Performance. }

We characterize \textit{FlexAttention} kernel performance by benchmarking across varying sequence lengths, attention variants, and numbers of heads, with kv size fixed at 256 MiB, head dimension at 64, and data type set to bfloat16. \par

\begin{figure*}[!ht]
    \centering
  \begin{subfigure}
         \centering
         \includegraphics[width=0.9\textwidth]{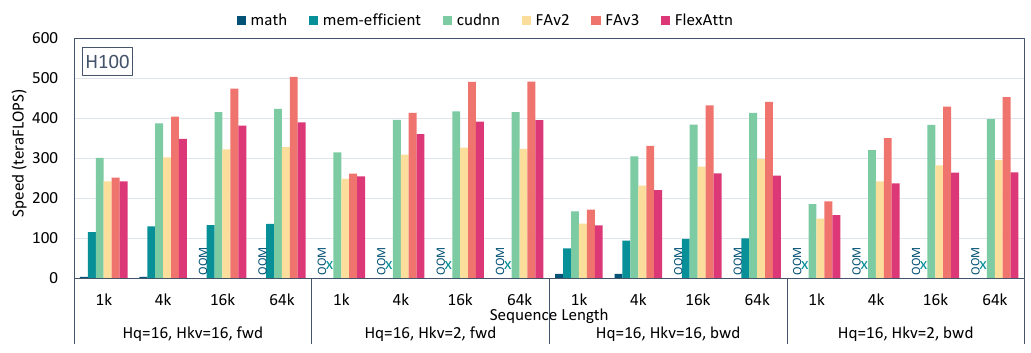}
    \end{subfigure}
     \begin{subfigure}
         \centering
         \includegraphics[width=\textwidth]{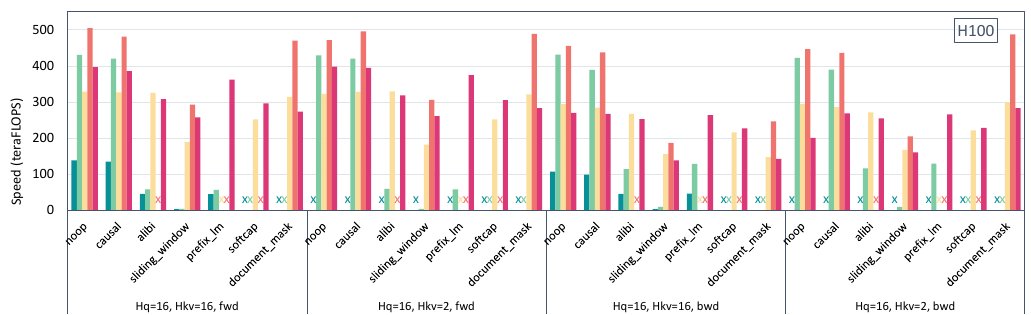}
    \end{subfigure}
    \caption{Attention Kernel Speed: Forward and Backward. \textbf{Top:} Causal Mask on QKV Length Ranging from 1k to 64k w/wo GQA. \textbf{Bottom: } Different Attention Variants on 16k-token-long QKV w/wo GQA.  }
    \label{fig:evaluation:benchmark_training}
\end{figure*}

\textbf{Training Performance.}  As shown in Figure \ref{fig:evaluation:benchmark_training}, assuming a causal mask, \textit{FlexAttention} yields a consistent 1.00x-1.22x speedup in the forward pass and 0.86x-1.05x speedup in the backward pass compared to {\tt FAv2} across different sequence lengths. For the 7 attention variants we evaluated, \textit{FlexAttention} achieves a 0.68x-1.43x speedup relative to {\tt FAv2} when {\tt FAv2} supports the variant. For the variants lacking native support, \textit{FlexAttention} achieves a 5.49x-8.00x speedup compared to SDPA kernels with itemized attention masks by computing the mask from {\tt mask\_mod} at runtime, thereby avoiding the need to realize and load the itemized mask. 

\begin{figure*}[!h]
    \centering
  \begin{subfigure}
         \centering
         \includegraphics[width=0.49\linewidth]{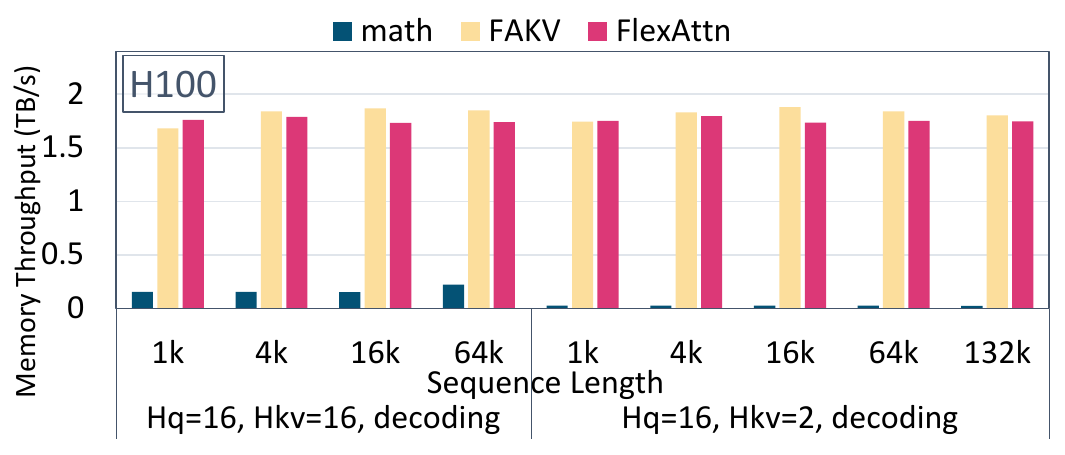}
    \end{subfigure}
     \hfill
     \begin{subfigure}
         \centering
         \includegraphics[width=0.49\linewidth]{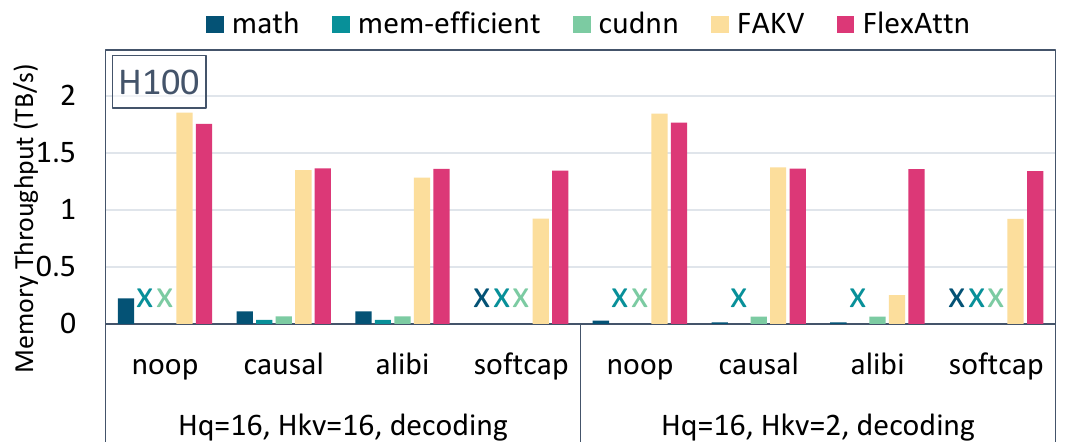}
    \end{subfigure}
    \caption{Attention Kernel Speed: Decoding for 1 Query Token. \textbf{Left:} Classic Attention on KV Length Ranging from 1k to 132k w/wo GQA. \textbf{Right: } Different Attention Variants on 16k-token-long KV cache w/wo GQA.  }
    \label{fig:evaluation:benchmark_decoding}
\end{figure*}
\par 
\textbf{Inference Performance. } As shown in Figure \ref{fig:evaluation:benchmark_decoding}, for a query length of 1, \textit{FlexAttention} delivers decoding performance comparable to FlashDecoding ({\tt FAKV}), achieving a 0.93x-1.45x speedup, except one outlier: \textbf{\textit{FlexAttention} is 5.37x faster than {\tt FAKV} when using GQA with {\tt alibi}}. This is an example of the "software lottery", where {\tt FAKV} lacks manual optimization for GQA with {\tt alibi}, and its fallback solution provides only 1/5 of the optimal performance. In contrast, \textit{FlexAttention} maintains consistent performance for this combination without manual tuning. 
Due to page limit, we leave Neighborhood Attention results in Appendix A.1.

\par
\begin{figure}[!h]
    \centering
    \includegraphics[width=0.9\linewidth]{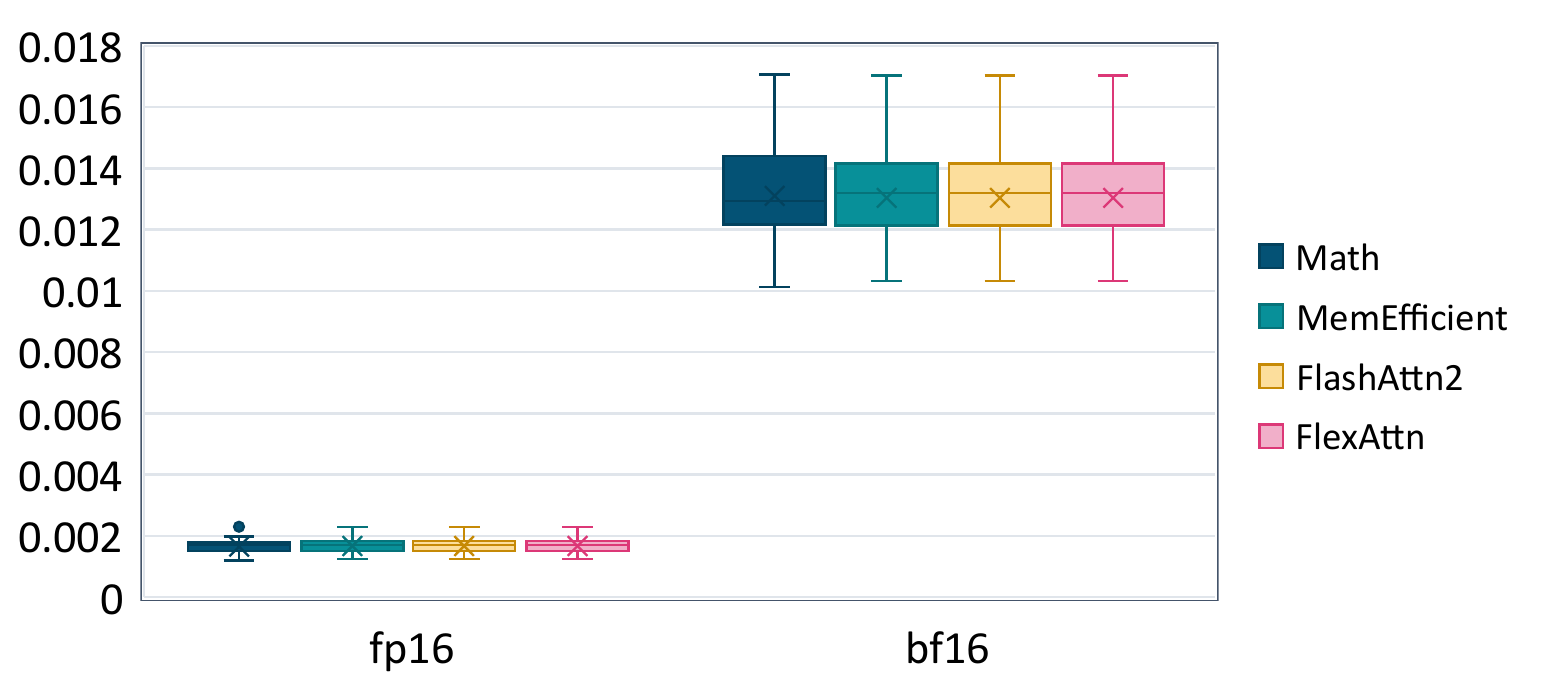}
    \caption{Root-mean-square error (RMSE) of bf16 and fp16 attention output compared to fp64 golden result. }
    \label{fig:evaluation:numerics}
\end{figure}
\textbf{Numeric Accuracy. } \textit{FlexAttention} does not introduce additional numeric errors compared to our baselines. (Figure \ref{fig:evaluation:numerics})

\subsection{End-to-end Performance. }
\textit{FlexAttention} boosts end-to-end training performance by over 2.4x and inference performance by up to 2.04x, and this speedup scales well with sequence length. We replace SDPA with \textit{FlexAttention} in {\tt gpt-fast} and {\tt torchtune} libraries and evaluate performance on LLaMa3 and LLaMa3.1 models. We show that \textit{FlexAttention} integrates well with the PyTorch framework in {\tt gpt-fast} and {\tt torchtune}, enabling optimizations such as CUDA graphs, parameter freezing, and kernel fusion, same as SDPA. \par

\begin{figure}[!h]
    \centering
    \includegraphics[width=\linewidth]{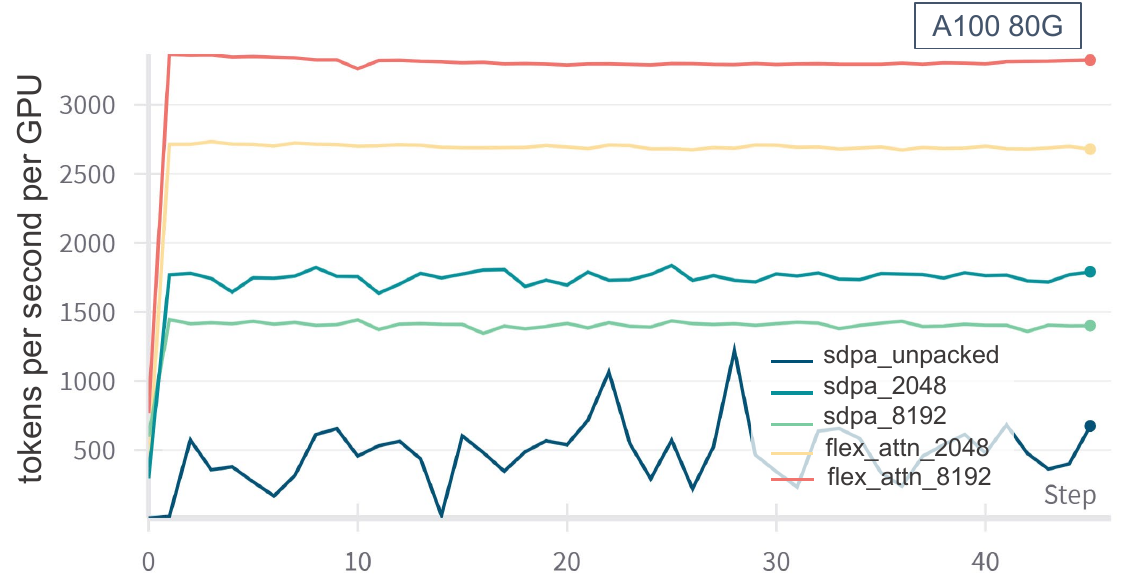}
    \caption{{\tt torchtune} Training Throughput on llama3-8B. }
    \label{fig:evaluation:torchtune}
\end{figure}

\textbf{Training Performance of {\tt torchtune}. } We set up {\tt torchtune} to fine-tune LLaMa3-8B  on the Alpaca~\cite{alpaca} dataset. To efficiently process the input sequences of different lengths, {\tt torchtune} concatenates them into jagged long sequences of fixed length. This approach requires a document mask that allows each input sequence to attend to itself while ignoring its neighbors. SDPA utilizes a precomputed boolean mask of size $B\times N \times N$, where B is batch size and N is the sequence length. As shown in Figure \ref{fig:evaluation:torchtune}, the cost of accessing this boolean mask grows quadratically, leading to a 25\% decrease in the training throughput when the sequence length increases from 2k to 8k. In contrast, \textit{FlexAttention} employs a {\tt BlockMask} and a document ID tensor of size $B \times N$ and  scales effectively with sequence length. \par

\begin{figure}[!h]
    \centering
    \begin{tabular}{@{}c@{}@{}c@{}}

    \includegraphics[width=0.51\linewidth]{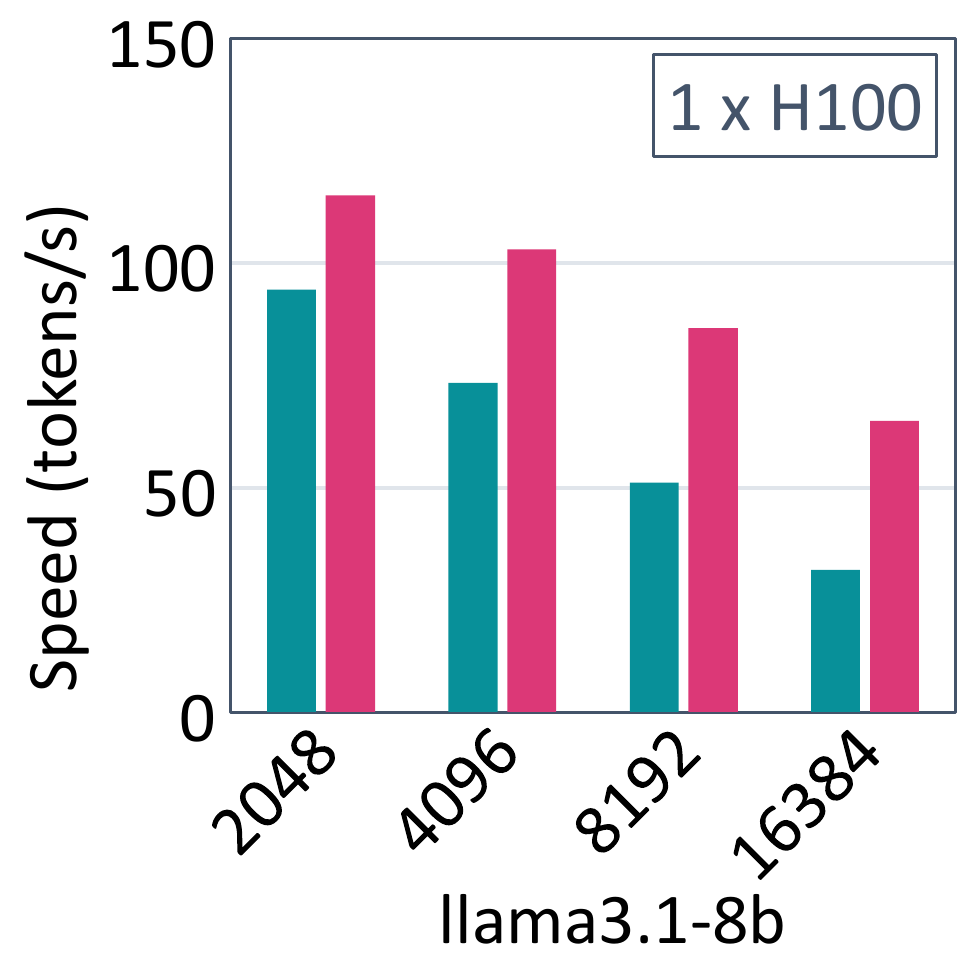} & 
    \includegraphics[width=0.48\linewidth]{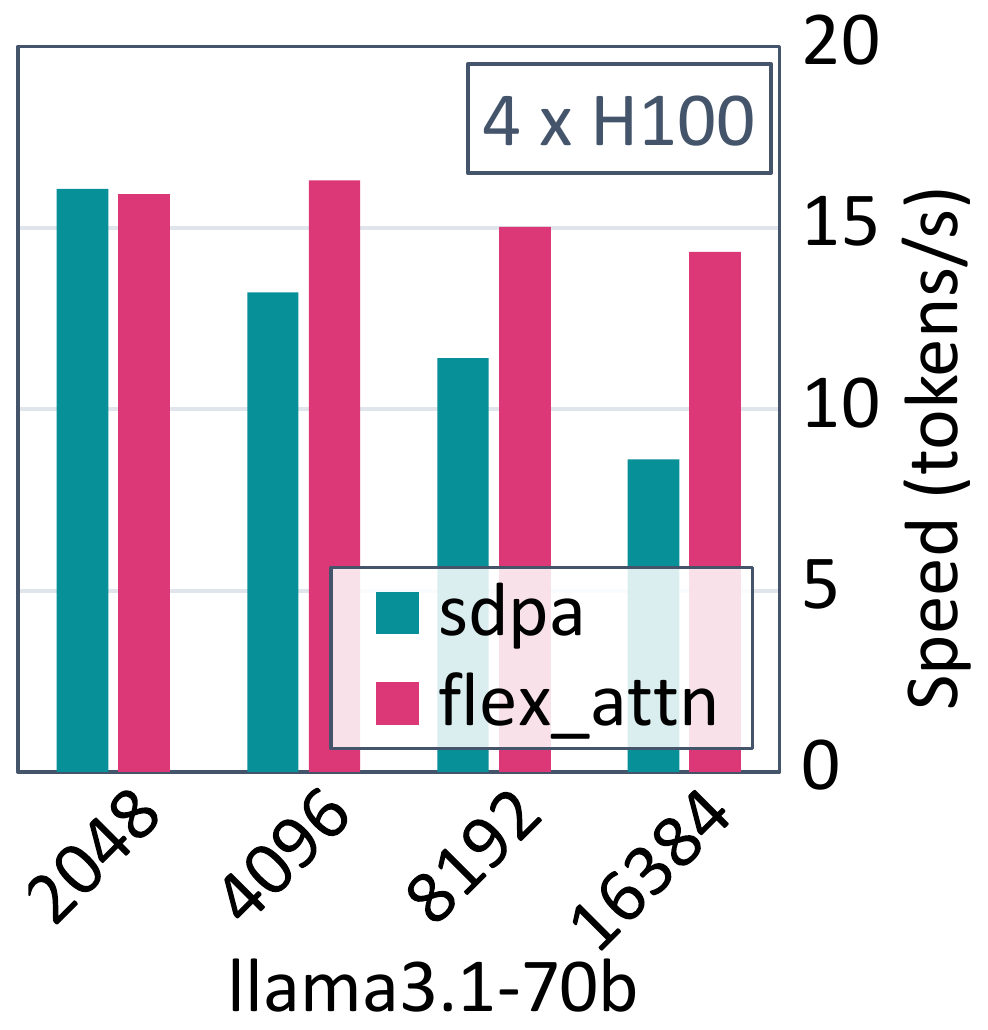} 
    \\
    \end{tabular}
    \caption{{\tt gpt-fast} Inference Speed on LLaMa3.1-8B and 70B.}
    \label{fig:evaluation:gpt-fast}
\end{figure}

\textbf{Inference Performance of {\tt gpt-fast}. } In Figure \ref{fig:evaluation:gpt-fast}, we show that \textit{FlexAttention} boosts LLaMa3.1-8B serving performance by 1.22x-2.04x, and LLaMa3.1-70B performance by 0.99x - 1.66x compared to SDPA. The speedup increases as context length grows and the attention kernel increasingly dominates the computation in each iteration. 

\subsection{Case Study: Paged Attention}

\begin{figure}[!h]
    \centering
  \begin{subfigure}
         \centering
         \includegraphics[width=\linewidth]{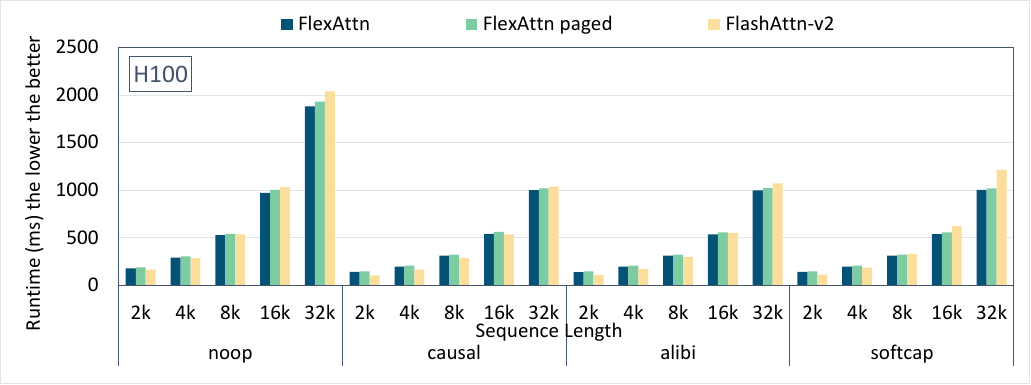}
    \end{subfigure}
    \hfill
    \begin{subfigure}
         \centering
         \includegraphics[width=\linewidth]{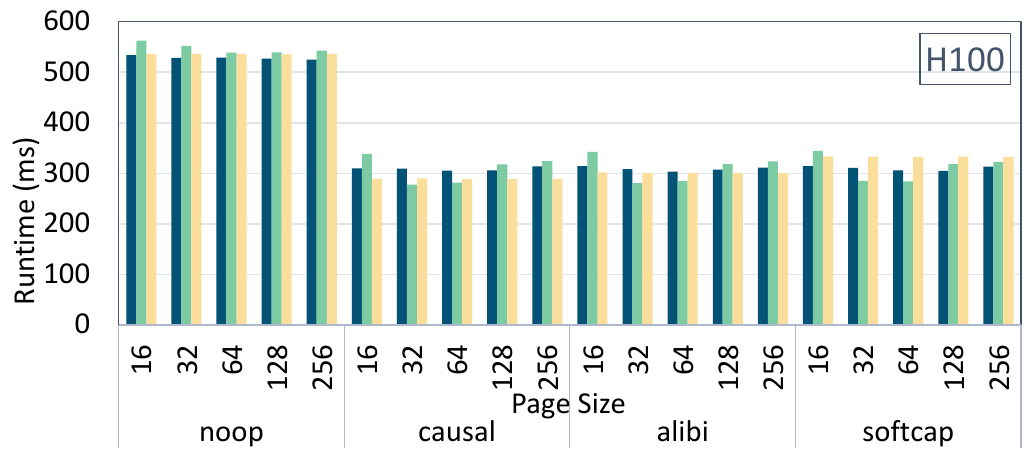}
    \end{subfigure}
    \caption{Runtime w/ and w/o paged attention (the lower the better). \textbf{Top:} Latency under diverse sequence length. \textbf{Bottom:} Latency under diverse page size.}
    \label{fig:evaluation:paged}
\end{figure}

While paged attention stores sentence requests of varying length in a compact physical KV cache, one major question is whether it would introduce high runtime overhead.
\autoref{fig:evaluation:paged}(a) shows runtime latency of flex attention with and without paged attention, as well as FlashAttn-v2.
We show runtime latency under changing sequence length, while keeping other dimension the same with batch size of 32, head dimension of 64, and number of heads as 16.
Overall, we observe less than 1\% runtime overhead on average when using \textit{FlexAttention} with paged attention, which is significantly smaller than the 20–26\% higher attention kernel overhead reported in vLLM~\cite{vLLM}.
The reason is that we do not introduce any kernel changes and rely on a fused indirect memory access to support \textit{FlexAttention} with paged attention.
Surprisingly, we even observe that \textit{FlexAttention} with paged attention is faster than FlashAttn-v2 without paged attention on large sequence lengths, which demonstrates the scalability of our design.

\autoref{fig:evaluation:paged}(b) demonstrates the impact from page size ranging from $16$ to $256$. Overall, we do not observe significant performance impact from changing page sizes.
Note that we manage physical KV cache in GPU global memory and do not swap memory to host disk, which mitigates the disk access overhead.

\section{Conclusion}
In this paper, we propose \textit{FlexAttention}, a programming model for generating optimized attention kernels.
While researchers continue to design new attention variants, they are often limited by the lack of hand-tuned kernels, leading to a large programmability and performance burden.
We hope that FlexAttention will allow researchers to explore new attention variants without being held back by what hand=written kernels support.


\bibliography{flex_att_ref}
\bibliographystyle{mlsys2024style/mlsys2024}

\clearpage
\appendix
\newpage

\paragraph{ }

\newpage

\section{Appendix}
\subsection{Neighborhood Attention (NA)}
\label{sec:na}
\begin{figure}[!h]
    \centering
    \begin{tabular}{@{}c@{}@{}c@{}}

    \includegraphics[width=0.37\linewidth]{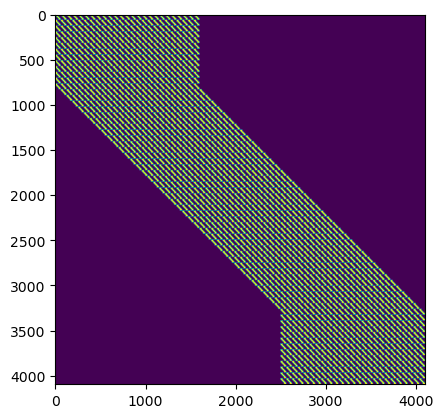} & 
    \includegraphics[width=0.3\linewidth]{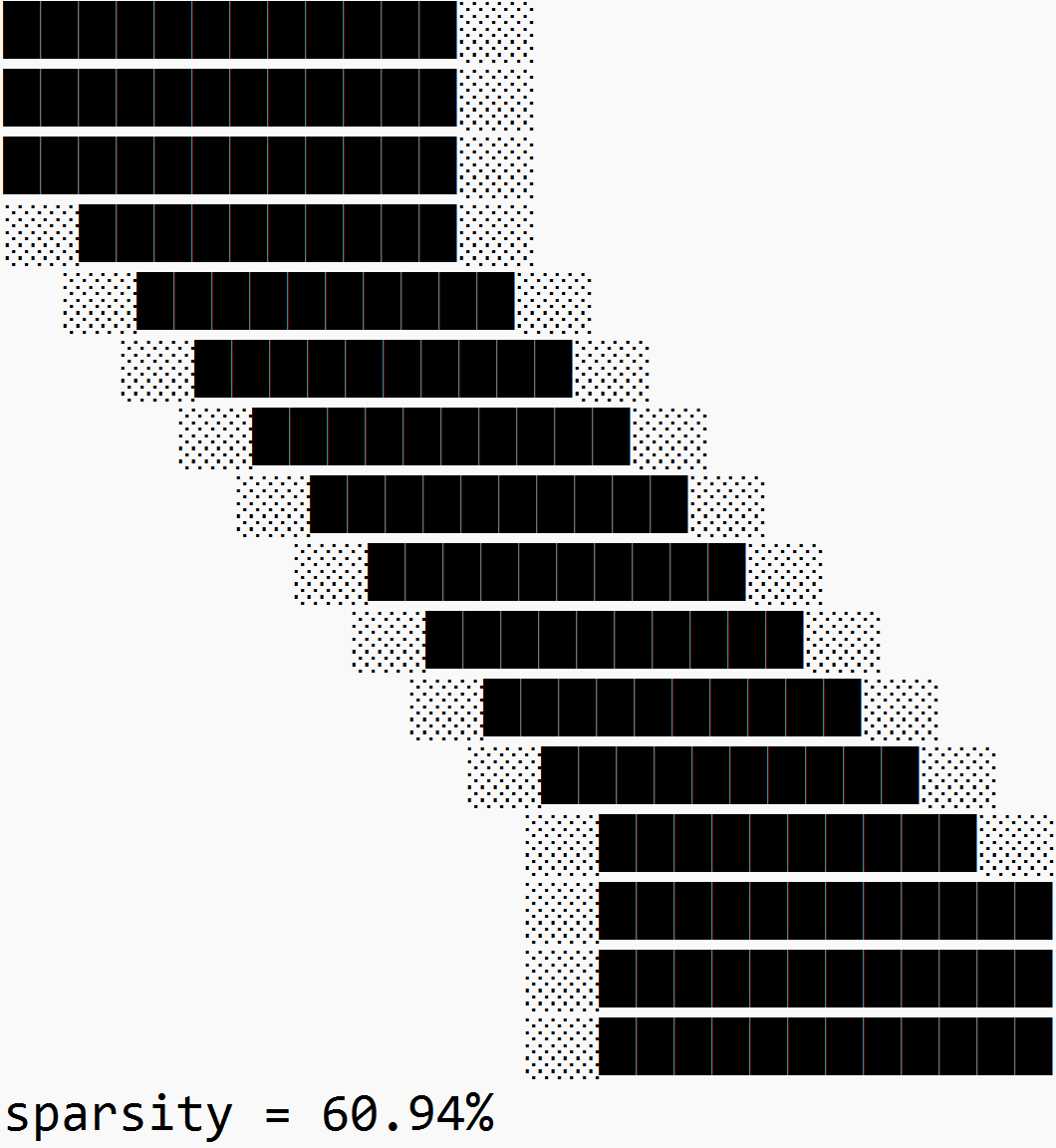} \\
    \includegraphics[width=0.37\linewidth]{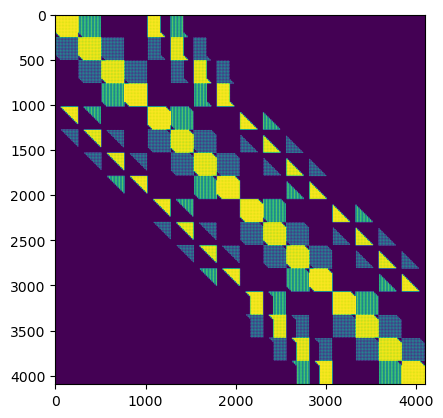} & 
    \includegraphics[width=0.3\linewidth]{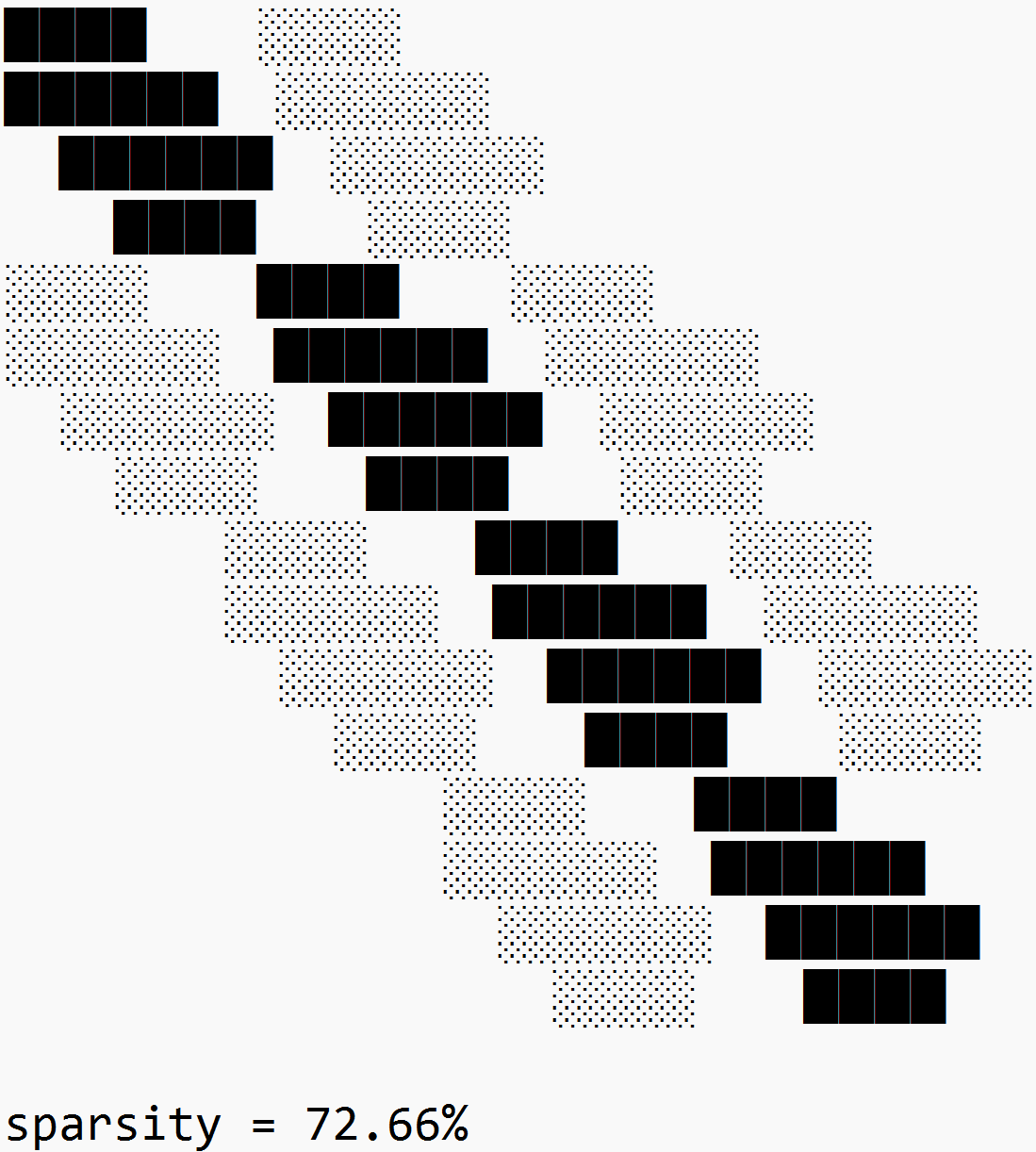} \\
    \includegraphics[width=0.37\linewidth]{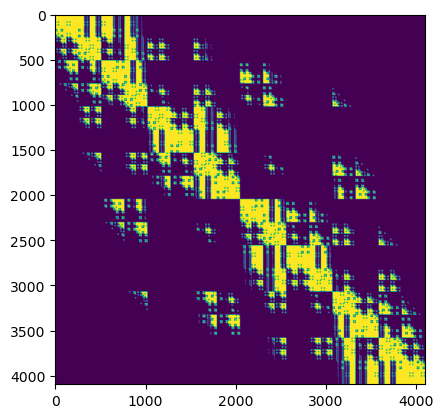} & 
    \includegraphics[width=0.3\linewidth]{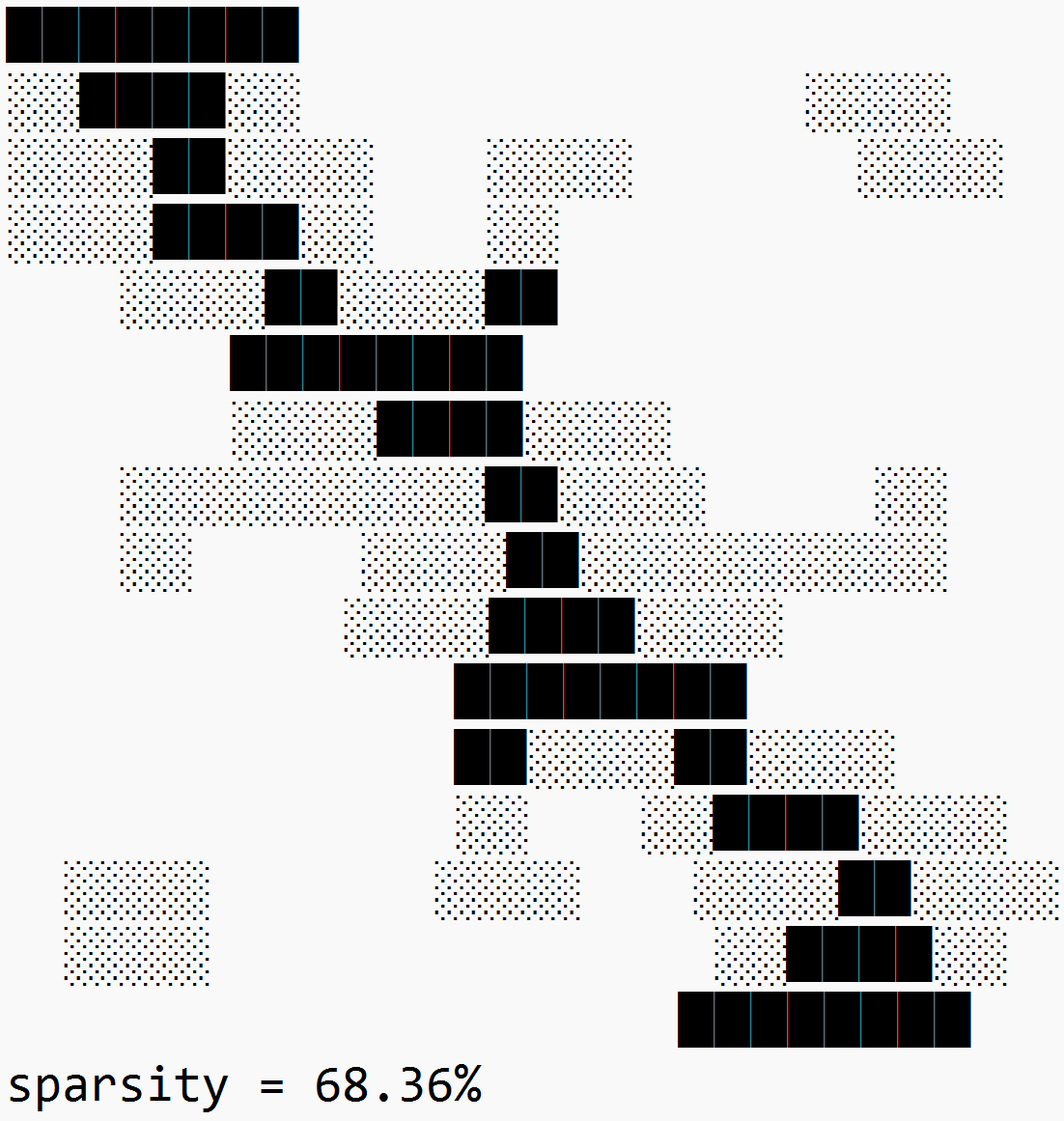} \\
  \end{tabular}
    \caption{NA Itemized Mask and the Corresponding Block Mask. \textbf{Top: } Naive NATTEN Mask. \textbf{Middle: } 2D-Tiled NA Mask. \textbf{Bottom: } Morton Curve NA Mask.}
    \label{fig:natten_masks}
\end{figure}

\begin{figure}[!h]
    \centering
  \begin{subfigure}
         \centering
         \includegraphics[width=\linewidth]{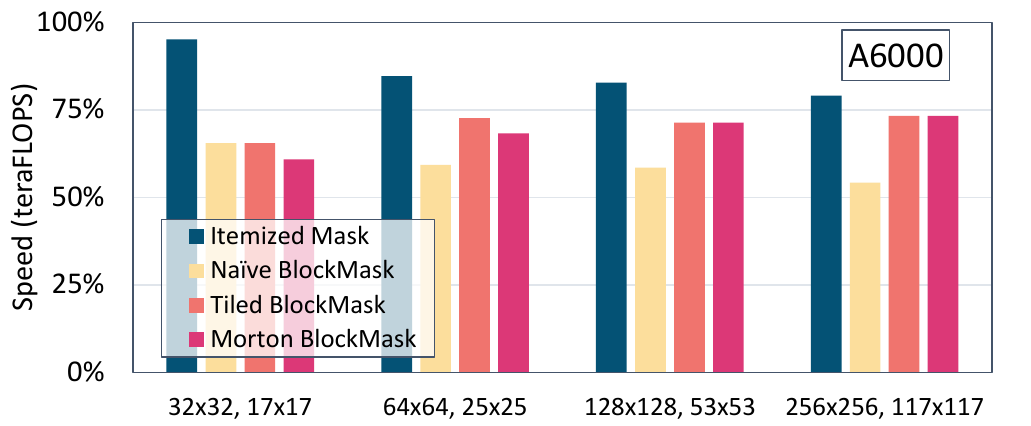}
    \end{subfigure}
    \hfill
    \begin{subfigure}
         \centering
         \includegraphics[width=\linewidth]{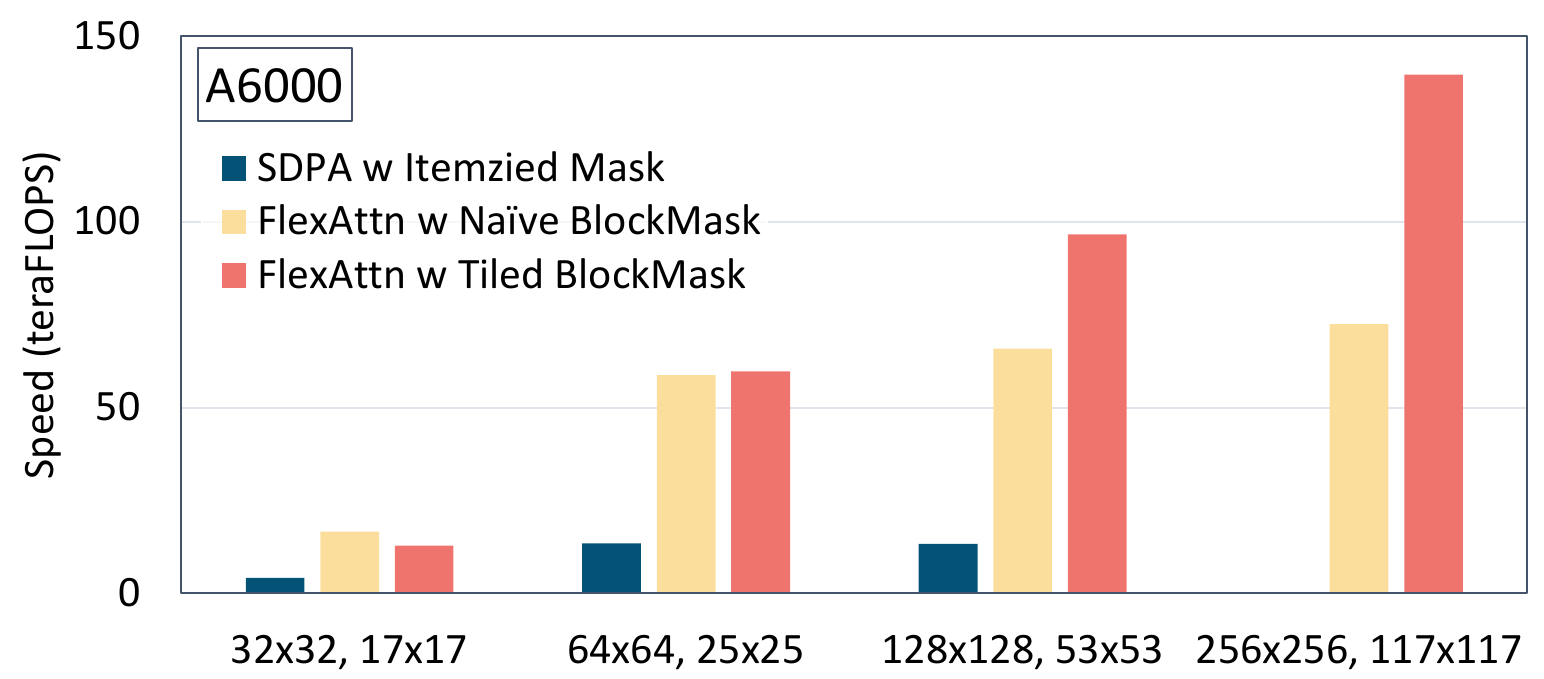}
    \end{subfigure}
    \caption{Neighborhood Attention with Different Mapping w.r.t canvas size, kernel size. \textbf{Top:} Mask Sparsity. \textbf{Bottom:} Speed.}
    \label{fig:evaluation:natten}
\end{figure}

Neighborhood Attention (NA)\cite{NAdeliated} is a local attention pattern used for 2D images, in which each pixel attends to its nearest neighboring pixels. 
NA mask is very complicated (Figure \ref{fig:natten_masks} top left) due to the 1D expansion of a 2D embedding neighborhood and thereby makes NA challenging to compute. 
Various advanced expansion strategies have been proposed\cite{NAtiled, NAfaster} to improve its block sparsity, but it is challenging manually to implement these optimizations in a high-performance attention kernel. 
Instead, we showcase that with \textit{FlexAttention}, the Tiled NA and Morton curve NA could be implemented in less than 10 lines of PyTorch code to exploit the sparsity of NA (Figure \ref{fig:natten_masks}) and enjoy its performance benefits (Figure \ref{fig:evaluation:natten}).

\end{document}